\documentclass[]{spie}  

\usepackage{amsmath,amsfonts,amssymb}
\usepackage{graphicx}
\usepackage[colorlinks=true, allcolors=blue]{hyperref}
\usepackage{CJKutf8}
\usepackage{subfiles}

\title{Analysis of LiDAR Configurations on Off-road Semantic Segmentation Performance}

\author[a]{Jinhee Yu}
\author[a, b]{Jingdao Chen}
\author[b]{Lalitha Dabbiru}
\author[b]{Christopher T. Goodin}

\affil[a]{Computer Science and Engineering, Mississippi State University, Mississippi State, United States}
\affil[b]{Center for Advanced Vehicular Systems, Mississippi State University, Mississippi State, United States}

\authorinfo{Further author information: (Send correspondence to Jinhee Yu)\\Jinhee Yu: E-mail: jy603@msstate.edu, Telephone: 662-325-7514}

\begin{document} 
\maketitle

\begin{abstract}

LiDAR-based 3D semantic segmentation is one of the most widely used perception methods to support scene understanding of self-driving vehicles. Most publicly available LiDAR datasets for driving scene segmentation, such as SemanticKITTI, nuScenes, and SemanticPOSS, provide only a single type of LiDAR configuration. Therefore, testing a trained model with a different channel configuration than the training dataset is sometimes inevitable in real-world applications. Despite the significance of this LiDAR channel mismatch problem in the machine learning pipeline, little research has focused on investigating the impact of the LiDAR configuration shift on a model's test performance. This paper aims to provide referenceable baseline experiments for the LiDAR configuration shifts. We explore the effect of using different LiDAR channels when training and testing a 3D LiDAR point cloud semantic segmentation model, utilizing Cylinder3D for the experiments. A Cylinder3D model is trained and tested on simulated 3D LiDAR point cloud datasets created using the Mississippi State University Autonomous Vehicle Simulator (MAVS) and 32, 64 channel 3D LiDAR point clouds of the RELLIS-3D dataset collected in a real-world off-road environment. Our experimental results demonstrate that sensor and spatial domain shifts significantly impact the performance of LiDAR-based semantic segmentation models. In the absence of spatial domain changes between training and testing, models trained and tested on the same sensor type generally exhibited better performance. Moreover, higher-resolution sensors showed improved performance compared to those with lower-resolution ones. However, results varied when spatial domain changes were present. In some cases, the advantage of a sensor's higher resolution led to better performance both with and without sensor domain shifts. In other instances, the higher resolution resulted in overfitting within a specific domain, causing a lack of generalization capability and decreased performance when tested on data with different sensor configurations.
\end{abstract}

\keywords{LiDAR, autonomous vehicles, semantic segmentation}

\section{Introduction}

The ability of autonomous vehicles to perceive their surrounding environment is crucial for identifying nearby objects and planning safe driving routes \cite{carballo2020}. Many autonomous driving systems depend on cameras and 3D Light Detection and Ranging (LiDAR) sensors to achieve this task. LiDAR sensors are particularly vital in off-road environments such as forests \cite{9551643}, construction sites \cite{kim2018crc}, and disaster sites \cite{chen2022aei} because they can compensate for the limitations of cameras by being able to directly measure distances and by being more robust to weather and lighting changes. Consequently, research on LiDAR semantic segmentation is being actively conducted to enable semantic scene understanding in off-road environments. LiDAR semantic segmentation involves assigning a label to each point in the point clouds collected from LiDAR sensors and classifying the corresponding points as objects or background elements. This process allows autonomous driving systems to recognize and differentiate the components of their surroundings, facilitating the planning of safe driving routes.

Despite the importance of LiDAR sensors in unstructured off-road environments, the number and diversity of open datasets featuring off-road LiDAR driving data are limited compared to those for on-road LiDAR driving datasets\cite{orfd2022,rs15010027,9551643}. Most annotated LiDAR driving point cloud datasets, such as SemanticKITTI \cite{behley2019semantickitti}, nuScenes\cite{caesar2020nuscenes}, and SemanticPOSS\cite{pan2020semanticposs}, primarily focus on on-road environments. Additionally, most of these datasets provide data collected using only a single LiDAR configuration. LiDAR configurations consist of various factors, such as field of view (FoV), range, resolution, and sensor mounting position. Changes in these factors can influence the accuracy and distribution of sensor data, potentially impacting the performance of semantic segmentation models. However, few studies have explored the effects of different LiDAR configurations on semantic segmentation performance.

Collecting and annotating off-road LiDAR point cloud datasets for autonomous driving requires significant time and effort. Consequently, it is sometimes necessary to use pre-trained models on existing datasets when conducting semantic segmentation in new off-road environments. However, obtaining an off-road LiDAR dataset that encompasses various off-road scenes and LiDAR configurations suitable for generalizing the pre-trained model's performance is challenging. Therefore, it is essential to understand how applying a pre-trained semantic segmentation model, trained on a specific driving dataset with a particular environment and configuration, affects its performance when tested on different off-road environments with varying LiDAR configurations.

In this paper, we aim to experimentally investigate the impact of sensor domain shifts caused by LiDAR configuration changes between training and testing sets on the performance of off-road semantic segmentation models. To achieve this, we conduct experiments using a real-world off-road LiDAR dataset from RELLIS-3D\cite{jiang2020rellis3d} and a simulated off-road LiDAR dataset generated using the Mississippi State University Autonomous Vehicle Simulator (MAVS)\cite{hudson2020mavs}. By training and testing a Cylinder3D\cite{zhu2021cylindrical} model for the task of semantic segmentation and examining the effect of different LiDAR configurations on the segmentation performance, we aim to address the following research questions:

\begin{enumerate}
        \item How does each configuration parameter change influence the semantic segmentation model's performance?
        \item What is the impact of sensor domain shift caused by configuration differences between training and testing data on the semantic segmentation model's performance?
        \item How does spatial domain shift (i.e., change in scene ecosystem) between training and testing data influence the semantic segmentation model's performance?
\end{enumerate}

This research aims to investigate the relationship between changes in LiDAR configurations and 3D LiDAR semantic segmentation in off-road environments, providing referenceable experimental results for other research and experiments. Furthermore, we expect that our findings will contribute to developing LiDAR datasets for autonomous driving systems in off-road environments and help improve and optimize LiDAR semantic segmentation model performance.

\label{sec:intro}  

\section{RELATED WORK}
\subsection{LiDAR technology in robotics}
\label{sec:title}

Light Detection and Ranging (LiDAR) technology enables an autonomous system to reconstruct the geometry of the surrounding environment in the form of a 3D point cloud.
LiDAR works by counting the time between acquiring backscattered energy from a pulsed laser beam and using these time measurements together with the speed of light in air to compute distances to objects and surfaces in the surrounding environment \cite{royo2019}. 
As the main form of 3D perception for many autonomous driving systems, LiDAR plays an important role in localization\cite{chen2016slam}, object classification \cite{chen2017pad}, scene segmentation\cite{chen2021ral}, and traversability estimation \cite{hirose2018gonet}.
Compared to cameras, LiDAR is advantageous for navigational tasks because it can be used to directly measure the distance to obstacles which enables robots to navigate effectively while avoiding obstacles \cite{chen2018ur}.
LiDAR also enables dynamic reconstruction of a robot's workspace \cite{price2020} which is useful for tracking and manipulating objects.

Primary LiDAR models used in robotics research include those manufactured by Velodyne, Ouster, Hesai, and RoboSense \cite{carballo2020}.
With the growth of research interest in autonomous driving, there is a wide variety of LiDAR models in the market with different hardware configurations and specifications such as (i) number of channels, which controls the number of LiDAR beams in the vertical direction, (ii) horizontal resolution, which controls the horizontal spacing between consecutive LiDAR returns, (iii) vertical resolution, which controls the vertical spacing between consecutive LiDAR returns, (iv) field-of-view, which specifies the minimum and maximum LiDAR beam angle, and (v) maximum range, which determines the maximum distance that can be measured by the LiDAR. Some previous works have investigated the effect of LiDAR configuration changes for specific variables such as sensor position \cite{Hu_2022_CVPR} and scanning style \cite{kim2017}. Other works have explored ways to overcome the configuration differences through reinforcement learning-based adaptation \cite{zhang2021} or solving for the optimal configuration \cite{mou2018}. However, there still exists a research gap in comprehensively analyzing the effect of all LiDAR configuration parameters, as well as in analyzing the effect of differing LiDAR configuration between training and testing.

\subsection{LiDAR semantic segmentation}

In the context of LiDAR point clouds, semantic segmentation refers to the task of assigning a semantic label to every 3D point in the point cloud.
Semantic segmentation can be carried out using various methods such as region growing \cite{chen2021ral}, or point-level classifiers trained using machine learning methods.
Most recent works in semantic segmentation rely on the latter method paired with deep neural networks trained on large datasets to learn high-level features from point cloud data.
Some neural network architectures process a point cloud directly as a unordered set \cite{qi2016pointnet, chen2018icra} and use pooling layers to combine information across the whole set. Other architectures pre-process the point cloud using a voxel\cite{Hu_2021_ICCV} or cylindrical \cite{zhu2021cylindrical} grid before applying convolutional layers to extract features. On the other hand, transformer-based architectures\cite{Zhao_2021_ICCV} use self-attention layers instead of convolutional layers to operate on the point cloud. There are also architectures that make use of context information from point clouds collected from multiple views over time \cite{chen2019ral, yajima2021isarc}. Despite the wealth of available methods for semantic segmentation, their robustness to different configurations of LiDAR hardware has not been well studied.

\subsection{LiDAR datasets}

Most datasets currently used for training autonomous driving models such as Semantic KITTI \cite{behley2019semantickitti}, nuScenes \cite{caesar2020nuscenes}, and SemanticPOSS \cite{pan2020semanticposs} feature on-road environments. On the other hand, off-road environments contain challenging features such as changing terrain types and vegetation with complex geometry. Only a few datasets exists for off-road environments, which are RELLIS-3D \cite{jiang2020rellis3d}, ORFD \cite{orfd2022}, and CaT \cite{cat2022}; and out of these, only RELLIS-3D offers labeled LiDAR data. Due to the challenge in annotating LiDAR datasets, some studies rely on simulations to generate labeled data. Simulated datasets offer an advantage compared to real datasets in that manual annotations are not needed to generate ground truth labels \cite{dabbiru2020}. Recent simulated datasets include LidarSim \cite{Manivasagam_2020_CVPR}, which utilizes a catalog of 3D static maps and 3D dynamic objects, and KITTI-CARLA\cite{deschaud2021}, which uses the CARLA simulator to simulate a vehicle with sensors identical to the KITTI dataset. Overall, even though these standard datasets and benchmark are widely used to train and test autonomous driving models, the significance of domain shift caused by differing LiDAR configurations is still unclear. Additionally, no studies have analyzed the impact of LiDAR configuration on off-road datasets.

\label{sec:related_work}

\section{METHODOLOGY}
In this study, we investigate the impact of various LiDAR configurations on the performance of off-road semantic segmentation models by training and testing the Cylinder3D model on both simulated and real-world datasets. The simulated dataset is generated using the Mississippi State University Autonomous Vehicle Simulator (MAVS), which can automatically assign ground truth semantic labels. The real-world dataset is sourced from the publicly available RELLIS-3D dataset, containing LiDAR point clouds from an off-road environment on Texas A\&M University's Rellis Campus. This dataset provides manually annotated labels based on synched image dataset.

Our experiments are divided into two parts. In the first part, we use MAVS to generate a simulated dataset with three distinct types of LiDAR sensors: OS-1, Velodyne Ultra Puck, and VLP-16. We also create different versions of the VLP-16 sensor in MAVS with varying configuration parameters such as field of view (FoV), range, resolution, and sensor position. These simulated datasets are separately trained and tested on the Cylinder3D model to evaluate the impact of changing each LiDAR configuration on semantic segmentation performance. Furthermore, we examine the combined effects of spatial and sensor domain shifts by testing the trained model on two separate test datasets: one from the same off-road environment as the training set and another from a different off-road environment. In the second part of our experiments, based on the results from the simulated dataset, we analyze the influence of sensor domain shifts on the semantic segmentation model's performance in real-world settings using the RELLIS-3D dataset, which comprises LiDAR point clouds from OS-1-64 (64 channels) and Velodyne Ultra Puck (32 channels) sensors.

In the following subsections, we outline detailed descriptions of the datasets used in our experiments, including the \hyperref[sec:rellis3d]{RELLIS-3D} dataset and off-road LiDAR simulation datasets generated using \hyperref[sec:mavs]{MAVS}. We also discuss the \hyperref[sec:cylinder3d]{Cylinder3D} model utilized to train and test these datasets and discuss the experimental details and processes.

\subsection{RELLIS-3D Dataset}
\label{sec:rellis3d}

The RELLIS-3D dataset is one of the few real-world off-road driving datasets containing LiDAR point clouds. The dataset was collected on Texas A\&M University's Rellis Campus using various sensors, such as LiDAR and RGB-D sensors, mounted on a Clearpath Warthog robot. Each raw sensor data frame, including images and 3D LiDAR point clouds, is synchronized. Additionally, the dataset provides pixel-wise image annotations as well as point-wise LiDAR annotations derived from the image annotations.

In this paper, we utilize only the data collected from the two LiDAR sensors available in the RELLIS-3D dataset, namely the 64-channel Ouster OS1-64 and the 32-channel Velodyne Ultra Puck. Both LiDAR sensors utilized a spinning frequency of 10 Hz during data collection. The OS1-64 sensor has a 45-degree vertical FoV, while the Ultra Puck has a 40-degree FoV. The relative position between the two sensors was fixed, with the OS1-64 mounted 7 cm above the Ultra Puck. These two LiDAR datasets contain 15 classes, including void, grass, tree, pole, water, vehicle, log, person, fence, bush, concrete, barrier, puddle, mud, and rubble. We excluded points in the dirt, asphalt, object, and building classes other than the 15 used by the authors when testing their dataset for LiDAR semantic segmentation models because those classes have only a few points\cite{jiang2020rellis3d}.

\begin{table}[h]
\caption{Details about the RELLIS-3D dataset with different splitting methods}
\label{tab:rellis-details}
\begin{center}
\begin{tabular}{l|c|c|c|c}
\hline
\rule[-1ex]{0pt}{3.5ex} & \multicolumn{2}{|c|}{Split Method 1 (original)} & \multicolumn{2}{|c}{Split Method 2 (modified)} \\ 
\hline
\rule[-1ex]{0pt}{3.5ex} Sensors & Velodyne Ultra Puck & OS1-64 & Velodyne Ultra Puck & OS1-64 \\
\hline
\hline
Number of Scans & & & & \\
\rule[-1ex]{0pt}{3.5ex} Train & 7792 & 7800 & 9296 & 9313 \\
\rule[-1ex]{0pt}{3.5ex} Valid & 2409 & 2413 & 2180 & 2184 \\
\rule[-1ex]{0pt}{3.5ex} Test & 3335 & 3343 & 2056 & 2059 \\
\hline
Number of Points & & & & \\
\rule[-1ex]{0pt}{3.5ex} Train & 191M & 576M & 224M & 726M \\
\rule[-1ex]{0pt}{3.5ex} Valid & 64M & 205M & 54M & 147M \\
\rule[-1ex]{0pt}{3.5ex} Test & 76M & 233M & 52M & 139M \\
\hline
Number of Classes Present & & & & \\
\rule[-1ex]{0pt}{3.5ex} Train & 14 & 14 & 14 & 14 \\
\rule[-1ex]{0pt}{3.5ex} Valid & 10 & 10 & 12 & 12 \\
\rule[-1ex]{0pt}{3.5ex} Test & 14 & 14 & 12 & 12 \\
\hline
\end{tabular}
\end{center}
\end{table} 

The data split is performed in two ways to examine the impact of LiDAR configuration changes on model performance with and without spatial domain shifts. The first approach follows the same procedure used by the authors in their published benchmark\cite{jiang2020rellis3d}, distributing all five off-road scenes collected from different locations evenly across the training, validation, and test sets. The second approach involves assigning scenes \textit{00}, \textit{01}, and \textit{02} as the training set, scene \textit{03} as the validation set, and scene \textit{04} as the test set. With the first approach, LiDAR scans from the same scene could appear in both the training and test sets. In contrast, the second approach ensures that the training and test sets comprise data collected from distinct locations. Table \ref{tab:rellis-details} illustrates the changes in the number of scans, points, and classes of the RELLIS-3D dataset according to each splitting method. Additionally, Appendix \ref{sec:miscB} provides point cloud class distribution graphs corresponding to each split approach.

\subsection{MAVS}
\label{sec:mavs}

Creating LiDAR datasets by changing a single sensor configuration parameter at a time while keeping other factors constant in the real world is practically impossible. Numerous factors, such as weather, lighting, and driving conditions, affect sensor performance, making it difficult to control these factors precisely when creating datasets with different sensor configurations. Furthermore, even when collecting data using multiple sensors simultaneously, each sensor must be mounted at different locations and operate at different times, necessitating synchronization. However, accurately synchronizing all sensor data is quite challenging. Moreover, accurately annotating all collected datasets, typically performed by humans, is time-consuming, labor-intensive, and error-prone. To conduct repeatable experiments in a precisely controlled environment without those problems and possibly erroneous annotations, we generate simulated datasets using MAVS (Mississippi State University Autonomous Vehicle Simulator). MAVS is an autonomous vehicle simulator developed by Mississippi State University, employed as a virtual simulation platform for developing and testing autonomous driving technologies in both on- and off-road environments. It facilitates the simulation and control of various virtual off-road scenes and automatically annotates sensor data from the simulated scenes.

\begin{table}[h]
\caption{Three different LiDAR sensor models used in the MAVS dataset. For the horizontal resolution values, the smaller the number, the higher the resolution.}
\label{tab:mavs_simulated_sensors}
\begin{center}
\begin{tabular}{l|c|c|c}
\hline
\rule[-1ex]{0pt}{3.5ex} Simulated Sensors & VLP-16 & Velodyne Ultra Puck & OS1-64 \\
\hline
\rule[-1ex]{0pt}{3.5ex} Maximum Range & 100m & 200m & 120m \\
\rule[-1ex]{0pt}{3.5ex} Vertical FoV & 30° (± 15°) & 40° (-25° - +15°) & 45° (± 22.5°) \\
\rule[-1ex]{0pt}{3.5ex} Vertical Resolution & 16 & 32 & 64 \\
\rule[-1ex]{0pt}{3.5ex} Horizontal FoV & 360° & 360° & 360° \\
\rule[-1ex]{0pt}{3.5ex} Horizontal Resolution & 0.2° & 0.2° & 0.18° \\
\rule[-1ex]{0pt}{3.5ex} Rotation Rate & 10Hz & 10Hz & 10Hz \\
\hline
\end{tabular}
\end{center}
\end{table} 

In this paper, we generate simulated datasets with five classes: rough trail, low vegetation, high vegetation, smooth trail, and obstacle. These datasets are collected under identical simulated environmental conditions using three different virtual LiDAR sensors configured as shown in Table \ref{tab:mavs_simulated_sensors}: OS1-64, Velodyne Ultra Puck (32 channels), and VLP-16. Furthermore, to investigate the impact of individual LiDAR configuration parameters such as maximum range, field of view (FoV), number of channels, horizontal resolution, and sensor position on semantic segmentation, we create twelve supplementary datasets using multiple VLP-16-based LiDAR configurations, modifying only one parameter at a time.
In our experiments, the default VLP-16 configuration is shown in Table \ref{tab:mavs_simulated_sensors}, featuring 16 channels, a maximum detection range of 100m, a vertical FoV between -15 and 15 degrees, a horizontal resolution of 0.2, and a sensor position of x = 0m, y = 0m, and z = 1.2m, relative to the center of the vehicle at the ground level. In the supplementary datasets, we introduce changes to the default configuration as follows: channels set to 32 or 64, maximum range adjusted to 150m or 200m, FoV maintained at 30 degrees while modifying the angles by 5 degrees up or down, horizontal resolution set to 0.1 or 0.4, and LiDAR position adjusted by 10 cm in each direction (up, down, left, and right). Each of the simulated datasets contains four scenes (\textit{00, 01, 02, 03}), with three scenes (\textit{00, 01, 02}) sharing the same simulated ecosystem and one scene (\textit{03}) using a different simulated ecosystem. The scenes contain 2702, 2717, 2717, and 2792 scans, respectively. Appendix~\ref{sec:miscA} provides further information about the differences.

\begin{figure} [ht]
   \begin{center}
   \begin{tabular}{c} 
   \includegraphics[height=2.9cm]{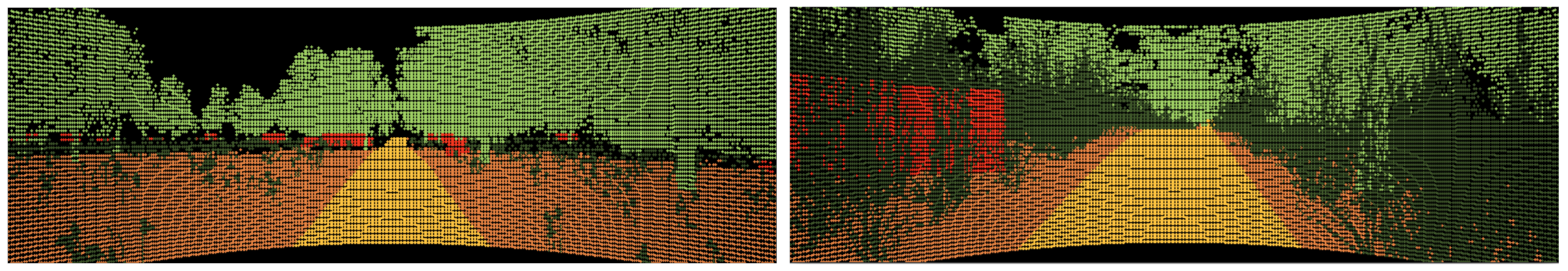}
	\end{tabular}
	\end{center}
   \caption[example] 
   { \label{fig:mavs-different-ecosystem} 
Example simulated off-road environments in MAVS with different ecosystems. Left: Labeled 3D LiDAR point cloud for off-road scenes \textit{00}, \textit{01}, and \textit{02}, which share the same ecosystem. Right: Labeled 3D LiDAR point cloud for off-road scene \textit{03}, which is from a different ecosystem} 
   \end{figure}

Similar to the RELLIS-3D dataset experiment, we apply two splitting approaches to the datasets created through MAVS to investigate the impact of LiDAR configuration shifts accompanied by spatial domain shifts. The first approach assigns three scenes with the same environmental ecosystem presets (which control factors such as tree species, vegetation height, and vegetation density) to the training (\textit{00}), validation (\textit{01}), and test sets (\textit{02}). This means that the training, validation, and test sets will have similar class distributions. The second approach uses two different types of scenes using different environmental ecosystem presets. The training (\textit{00}) and validation (\textit{01}) sets use the first scene type, whereas the test set uses a different scene type (\textit{03}). This means the training and test sets will have different vegetation species in their off-road scenes. Figure \ref{fig:mavs-different-ecosystem} illustrates the differences in datasets collected under two different simulated off-road scene conditions, even though the same virtual sensor is used. 

\subsection{Cylinder3D}
\label{sec:cylinder3d}

Our study employs the Cylinder3D model, a neural network for LiDAR semantic segmentation in driving scenarios proposed by Zhu et al. \cite{zhu2021cylindrical}. This model leverages point clouds' 3D topology relationships and structures in driving scenes, effectively capturing local and global contextual information to enhance segmentation performance. Cylinder3D consists of several key components, including cylinder partition, asymmetric residual blocks, and dimension-decomposition-based context modeling. Cylinder3D adopts a cylindrical representation to improve computational efficiency and accuracy by converting LiDAR point clouds into a cylinder coordinate system. This representation allows for a more even distribution of points across different regions while preserving the geometric structure.

For the simulated datasets in MAVS, Cylinder3D is trained for 6 epochs, with a batch size of 2 and a learning rate of 0.001. For the RELLIS-3D dataset, Cylinder3D is trained for 40 epochs with a batch size of 2 and a learning rate of 0.001.

At the time of its release, Cylinder3D was the best-performing semantic segmentation model on the  SemanticKITTI\cite{behley2019semantickitti} and nuScenes\cite{caesar2020nuscenes} datasets. At the time of this study, Cylinder3D placed 5th on the  SemanticKITTI leaderboard and 10th  on the nuScenes leaderboard. Due to its availability of open-source code and good performance in 3D semantic segmentation for outdoor scenes, we chose to use Cylinder3D for our experiments.
\label{sec:methodology}

\section{RESULTS}

\subsection{Evaluation Metrics}

This study evaluates the Cylinder3D model's performance using the mean Intersection over Union (mIoU) metric. mIoU is a widely used metric for semantic segmentation \cite{behley2019semantickitti, caesar2020nuscenes}, as it quantifies the overlap between the predicted segmentation and the ground truth. A higher mIoU value indicates better model performance.

The mIoU is calculated as follows:

\begin{equation} mIoU = \frac{1}{N} \sum_{i=1}^{N} \frac{TP_i}{(TP_i + FP_i + FN_i)} \end{equation}

where $N$ is the number of classes, $TP_i$ is the number of true positive predictions for class $i$, $FP_i$ is the number of false positive predictions for class $i$, and $FN_i$ is the number of false negative predictions for class $i$. In our experiments, we compute the mIoU for each class and average them to obtain the final mIoU value.

\subsection{Effect of Change in Individual LiDAR Parameter}
\label{sub-sec:Effect of Change in Individual LiDAR Parameter}
   \begin{figure} [ht]
   \begin{center}
   \resizebox{\textwidth}{!}{%
   \begin{tabular}{c} 
   \includegraphics{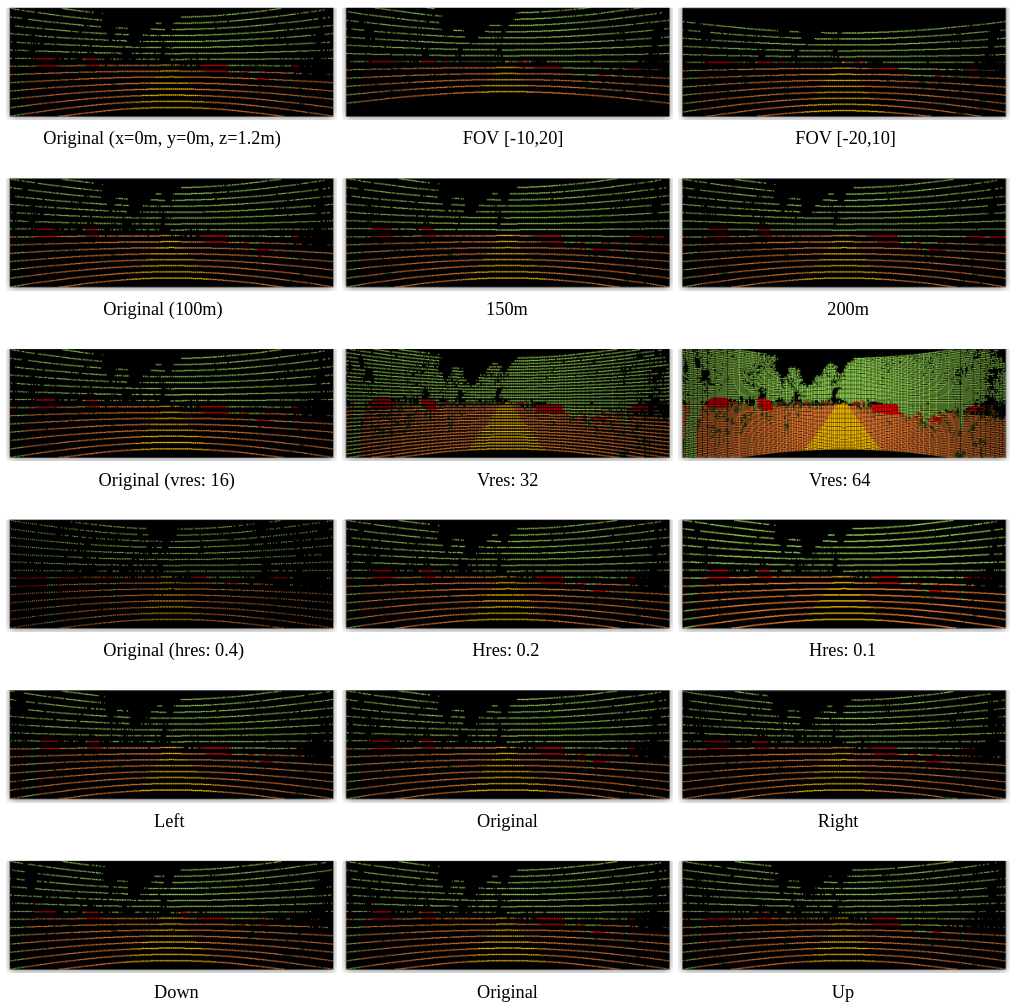}
	\end{tabular}} 
	\end{center}
   \caption[example] 
   { \label{fig:mavs-visual} 
Visualizations of simulated datasets generated using MAVS, comparing the effects of different LiDAR configuration parameters on point cloud data. The top row shows point clouds generated with varying field of view (FoV), the second row shows variations in range values, and the third row shows the impact of changing the vertical resolution. The fourth row shows the effect of modifying the horizontal resolution, while the bottom two rows show the impact of adjusting the LiDAR position. } 
   \end{figure} 

To establish a baseline for the Cylinder3D model performance without sensor configuration changes between training and testing, we trained the model on MAVS simulated datasets, with single sensor configuration parameters altered. Figure \ref{fig:mavs-visual} illustrates visualizations of the generated simulated datasets using MAVS, comparing the visual changes caused by the different LiDAR configuration parameters in the point clouds. The first experiment investigated the impact of altering individual LiDAR parameters without spatial domain shift (Table \ref{tab:mavs-same-indiv-results}). We observed that the model performance remained relatively stable when the range was adjusted from 100m to 150m or 200m, with mIoU decreasing only by 0.2\% and 0.3\%, respectively. The mIoU fluctuated as the field of view (FoV) was modified; the model performed better with a [-20, 10] FoV configuration than the original [-15, 15] FoV, increasing by 0.25\% mIoU. However, the mIoU decreased by 1.5\% when the FoV was set to [-10, 20]. A consistent improvement in the model's performance was observed as the vertical resolution increased, with mIoU increasing by 1.62\% when the vertical resolution was doubled from 16 to 32 and by 2.26\% when it was quadrupled to 64. The model's performance improved with an increase in horizontal resolution from 0.2 to 0.1, resulting in a 1.42\% increase in mIoU, while a decrease in resolution to 0.4 resulted in a 1.55\% reduction in mIoU. The model's performance slightly varied with changes in LiDAR position; when the LiDAR was moved 10 cm up, down, left, or right, the mIoU fluctuated between 93.92\% and 94.71\%. The most significant decrease in mIoU occurred when the LiDAR was moved 10 cm to the right, resulting in a decline of 0.7\%. Overall, the model appeared relatively robust to small changes in LiDAR position in this scenario. In the absence of sensor and spatial domain shifts, the Cylinder3D model was generally robust to changes in LiDAR parameters, with some sensitivity to the field of view and horizontal resolution.

\begin{table}[htbp!]
\caption{Performance metrics for the Cylinder3D model on MAVS simulated datasets with varying individual LiDAR parameters under no spatial domain shift.}
\label{tab:mavs-same-indiv-results}
\begin{center}
\resizebox{\textwidth}{!}{%
\begin{tabular}{l|c|c|c|c|c|c}
\hline
\rule[-1ex]{0pt}{3.5ex} Configuration & High Vegetation & Low Vegetation & Obstacle & Rough Trail & Smooth Trail & Mean \\
\hline
\hline
\rule[-1ex]{0pt}{3.5ex} \textbf{Modified maximum range} & & & & & & \\
\rule[-1ex]{0pt}{3.5ex} 100m (original) & 99.31 & 90.36 & 92.46 & 98.86 & 92.12 & 94.62 \\
\rule[-1ex]{0pt}{3.5ex} 150m & 99.11 & 90.91 & 90.28 & 98.74 & 93.17 & 94.44 \\
\rule[-1ex]{0pt}{3.5ex} 200m & 99.14 & 90.83 & 90.32 & 98.56 & 92.77 & 94.32 \\
\hline
\rule[-1ex]{0pt}{3.5ex} \textbf{Modified FoV} & & & & & & \\
\rule[-1ex]{0pt}{3.5ex} [-15, 15] (original) & 99.31 & 90.36 & 92.46 & 98.86 & 92.12 & 94.62 \\
\rule[-1ex]{0pt}{3.5ex} [-10, 20] & 99.21 & 91.47 & 88.87 & 98.32 & 87.71 & 93.12 \\
\rule[-1ex]{0pt}{3.5ex} [-20, 10] & 98.83 & 92.48 & 90.12 & 98.76 & 94.18 & 94.87 \\
\hline
\rule[-1ex]{0pt}{3.5ex} \textbf{Modified vertical resolution} & & & & & & \\
\rule[-1ex]{0pt}{3.5ex} (vres: 16) (original) & 99.31 & 90.36 & 92.46 & 98.86 & 92.12 & 94.62 \\
\rule[-1ex]{0pt}{3.5ex} (vres: 32) & 99.54 & 94.70 & 93.98 & 98.98 & 93.98 & 96.24 \\
\rule[-1ex]{0pt}{3.5ex} (vres: 64) & 99.69 & 95.15 & 95.37 & 98.99 & 95.20 & 96.88 \\
\hline
\rule[-1ex]{0pt}{3.5ex} \textbf{Modified horizontal resolution} & & & & & & \\
\rule[-1ex]{0pt}{3.5ex} (hres: 0.1) & 99.44 & 92.61 & 94.51 & 99.21 & 94.42 & 96.04 \\
\rule[-1ex]{0pt}{3.5ex} (hres: 0.2) (original) & 99.31 & 90.36 & 92.46 & 98.86 & 92.12 & 94.62 \\
\rule[-1ex]{0pt}{3.5ex} (hres: 0.4) & 99.00 & 87.14 & 88.71 & 98.77 & 91.73 & 93.07 \\
\hline
\rule[-1ex]{0pt}{3.5ex} \textbf{Modified position} & & & & & & \\
\rule[-1ex]{0pt}{3.5ex} (0m, 0m, 1.2m) (original) & 99.31 & 90.36 & 92.46 & 98.86 & 92.12 & 94.62 \\
\rule[-1ex]{0pt}{3.5ex} 10 cm up & 99.33 & 90.84 & 91.10 & 98.94 & 91.96 & 94.433 \\
\rule[-1ex]{0pt}{3.5ex} 10 cm down & 99.21 & 90.72 & 91.76 & 98.67 & 92.39 & 94.55 \\
\rule[-1ex]{0pt}{3.5ex} 10 cm left & 99.26 & 90.48 & 92.27 & 98.93 & 92.62 & 94.71 \\
\rule[-1ex]{0pt}{3.5ex} 10 cm right & 99.27 & 90.06 & 91.66 & 98.58 & 90.08 & 93.92 \\
\hline
\end{tabular}}
\end{center}
\end{table} 

In the second experiment, we examined the impact of altering individual LiDAR parameters when a spatial domain shift was present (Table \ref{tab:mavs-diff-indiv-results}). When the range was adjusted from 100m to 150m or 200m, the mIoU increased by 0.3\% when the range changed to 150m and by 3.43\% when it changed to 200m. This contrasted with the findings of the first experiment, which displayed a slight but consistent decrease in mIoU as the range increased. The influence of modifying the FoV was more significant than in the first experiment, with the mIoU decreasing by 12.99\% when the FoV was set to [-10, 20] and slightly increasing by 2.82\% when the FoV was set to [-20, 10]. Enhancing the vertical resolution positively impacted the model's performance, with the mIoU increasing by 3.56\% when the vertical resolution was doubled from 16 to 32 and by 3.58\% when it quadrupled to 64. In line with the first experiment's results, the model could benefit from higher vertical resolution when spatial domain shifts were present. The effect of altering the horizontal resolution was more noticeable when there was a spatial domain shift. When the horizontal resolution was increased from 0.2 to 0.1, the mIoU increased by 2.24\%, while the mIoU decreased by 4.41\% when the resolution was reduced to 0.4. The model's performance was more sensitive to changes in LiDAR position when a spatial domain shift was present. When the LiDAR was moved 10 cm up, down, left, or right, the mIoU fluctuated between 7.79\% and 44.51\%. The most substantial increases in mIoU occurred when the LiDAR was moved 10 cm left and right, resulting in a decrease of 30.37\% and 30.81\%, respectively. This indicates that the model is more sensitive to changes in LiDAR position when there is a spatial domain shift.

\begin{table}[h]
\caption{Performance metrics for the Cylinder3D model on MAVS simulated datasets with varying individual LiDAR parameters under spatial domain shift.}
\label{tab:mavs-diff-indiv-results}
\begin{center}
\resizebox{\textwidth}{!}{%
\begin{tabular}{l|c|c|c|c|c|c}
\hline
\rule[-1ex]{0pt}{3.5ex} Configuration & High Vegetation & Low Vegetation & Obstacle & Rough Trail & Smooth Trail & Mean \\
\hline
\hline
\rule[-1ex]{0pt}{3.5ex} \textbf{Modified maximum range} & & & & & & \\
\rule[-1ex]{0pt}{3.5ex} 100m (original) & 47.10 & 11.92 & 8.42  & 43.88 & 81.68 & 38.60 \\
\rule[-1ex]{0pt}{3.5ex} 150m & 48.62 & 13.95 & 3.48 & 49.51 & 78.94 & 38.90 \\
\rule[-1ex]{0pt}{3.5ex} 200m & 59.20 & 13.90 & 2.17 & 55.15 & 79.72 & 42.03 \\
\hline
\rule[-1ex]{0pt}{3.5ex} \textbf{Modified FoV} & & & & & & \\
\rule[-1ex]{0pt}{3.5ex} [-15, 15] & 47.10 & 11.92 & 8.42 & 43.88 & 81.68 & 38.60 \\
\rule[-1ex]{0pt}{3.5ex} [-10, 20] & 48.91 & 8.18  & 11.34 & 6.85  & 52.77 & 25.61 \\
\rule[-1ex]{0pt}{3.5ex} [-20, 10] & 41.58 & 10.29 & 2.67  & 67.91 & 84.67 & 41.42 \\
\hline
\rule[-1ex]{0pt}{3.5ex} \textbf{Modified vertical resolution} & & & & & & \\
\rule[-1ex]{0pt}{3.5ex} (vres: 16) & 47.10 & 11.92 & 8.42 & 43.88 & 81.68 & 38.60 \\
\rule[-1ex]{0pt}{3.5ex} (vres: 32) & 50.62 & 7.48  & 2.88 & 62.61 & 87.22 & 42.16 \\
\rule[-1ex]{0pt}{3.5ex} (vres: 64) & 48.99 & 4.91  & 2.88 & 68.74 & 85.38 & 42.18 \\
\hline
\rule[-1ex]{0pt}{3.5ex} \textbf{Modified horizontal resolution} & & & & & & \\
\rule[-1ex]{0pt}{3.5ex} (hres: 0.1) & 46.65 & 10.93 & 9.14 & 49.70 & 80.89 & 39.46 \\
\rule[-1ex]{0pt}{3.5ex} (hres: 0.2) (original) & 47.10 & 11.92 & 8.42  & 43.88 & 81.68 & 38.60 \\
\rule[-1ex]{0pt}{3.5ex} (hres: 0.4) & 57.96 & 11.81 & 4.93  & 60.11 & 71.29 & 41.22 \\
\hline
\rule[-1ex]{0pt}{3.5ex} \textbf{Modified position} & & & & & & \\
\rule[-1ex]{0pt}{3.5ex} (0m, 0m, 1.2m) (original) & 47.10 & 11.92 & 8.42  & 43.88 & 81.68 & 38.60 \\
\rule[-1ex]{0pt}{3.5ex} 10 cm up & 48.84 & 6.12  & 7.66 & 43.17 & 74.51 & 36.06 \\
\rule[-1ex]{0pt}{3.5ex} 10 cm down & 56.96 & 15.45 & 4.95 & 66.63 & 78.55 & 44.51 \\
\rule[-1ex]{0pt}{3.5ex} 10 cm left & 31.52 & 8.74  & 0.55 & 0.03  & 0.31  & 8.23  \\
\rule[-1ex]{0pt}{3.5ex} 10 cm right & 27.76 & 0.32  & 1.72 & 6.39  & 2.74  & 7.79  \\
\hline
\end{tabular}}
\end{center}
\end{table} 

In conclusion, the Cylinder3D model generally performed well even when individual LiDAR parameters were changed, provided there was no spatial domain shift. The experimental results demonstrated that, without a spatial domain shift, the Cylinder3D model is relatively robust to changes in LiDAR parameters, displaying a tendency for performance improvement when increasing vertical and horizontal resolution. Additionally, performance increased when the FoV was oriented more toward the ground. However, when a spatial domain shift occurred, the model became more sensitive to changes in specific parameters, such as field of view and horizontal resolution. In this scenario, an increase in vertical resolution and a lower FoV still contributed to performance improvement, but unlike the experiment without a spatial domain shift, increasing the horizontal resolution led to the degradation of the model's performance. Furthermore, sensitivity to LiDAR position changes increased, especially when moving left or right, resulting in substantial performance decreases.


\subsection{Combined Effects of Multiple LiDAR Configuration Parameters}
\label{sub-sec:Combined Effects of Multiple LiDAR Configuration Parameters}

We investigated the combined effects of multiple LiDAR configuration parameters in these experiments by performing tests on the MAVS simulated datasets generated with VLP-16, Velodyne Ultra Puck, and OS1 configurations (Table \ref{tab:mavs-same-combined-results}). It is important to note that each LiDAR has a different range, FoV, vertical, and horizontal resolution (described in Table \ref{tab:mavs_simulated_sensors}). A close examination of Table \ref{tab:mavs-same-combined-results} reveals that the Velodyne Ultra Puck configuration achieved the highest mean mIoU (96.25) across all classes, closely followed by the OS1-64 configuration (95.81). The VLP-16 configuration, on the other hand, had a slightly lower mean mIoU of 94.62.

\begin{table}[h]
\caption{Results of the experiments on the MAVS simulated datasets generated with VLP-16, Velodyne Ultra Puck, and OS1 configurations under no spatial domain shift.}
\label{tab:mavs-same-combined-results}
\begin{center}
\resizebox{\textwidth}{!}{%
\begin{tabular}{l|c|c|c|c|c|c}
\hline
\rule[-1ex]{0pt}{3.5ex} Configuration & High Vegetation & Low Vegetation & Obstacle & Rough Trail & Smooth Trail & Mean \\
\hline
\hline
\rule[-1ex]{0pt}{3.5ex} VLP-16 & 99.31 & 90.36 & 92.46 & 98.86 & 92.12 & 94.62 \\
\rule[-1ex]{0pt}{3.5ex} Velodyne Ultra Puck & 99.45 & 94.05 & 92.94 & 98.94 & 95.86 & 96.25 \\
\rule[-1ex]{0pt}{3.5ex} OS1-64 & 99.48 & 93.18 & 93.27 & 98.80 & 94.33 & 95.81 \\
\hline
\end{tabular}}
\end{center}
\end{table} 

Table \ref{tab:mavs-diff-combined-results} displays the results of the experiments under spatial domain shifts. The Velodyne Ultra Puck configuration again demonstrated the best overall performance with a mean mIoU of 46.16. The OS1-64 configuration followed with a mean mIoU of 44.32, and the VLP-16 configuration had the lowest mean mIoU at 38.60. The mean performance of the models across different LiDAR configurations was relatively high in the no spatial domain shift scenario, ranging from 94.62 to 96.25. Additionally, the standard deviations were low, indicating that the model's performance is consistent across different classes. In contrast, in the spatial domain shift scenario, the mean performance of the models was significantly lower, ranging from 38.60 to 46.16. The model's performance also varied considerably across different classes. This highlights the challenges faced by the model when generalizing to new environments.

\begin{table}[h]
\caption{Results of the experiments on the MAVS simulated datasets generated with VLP-16, Velodyne Ultra Puck, and OS1 configurations under spatial domain shift.}
\label{tab:mavs-diff-combined-results}
\begin{center}
\resizebox{\textwidth}{!}{%
\begin{tabular}{l|c|c|c|c|c|c}
\hline
\rule[-1ex]{0pt}{3.5ex} Configuration & High Vegetation & Low Vegetation & Obstacle & Rough Trail & Smooth Trail & Mean \\
\hline
\hline
\rule[-1ex]{0pt}{3.5ex} VLP-16 & 47.10 & 11.92 & 8.42 & 43.88 & 81.68 & 38.60 \\
\rule[-1ex]{0pt}{3.5ex} Velodyne Ultra Puck & 56.39 & 7.90  & 2.06 & 75.59 & 88.88 & 46.16 \\
\rule[-1ex]{0pt}{3.5ex} OS1-64 & 52.88 & 6.89  & 2.87 & 71.49 & 87.48 & 44.32 \\
\hline
\end{tabular}}
\end{center}
\end{table} 

Based on the two experimental results, we can see that the Velodyne Ultra Puck configuration consistently outperformed the other configurations with or without spatial domain shifts in these experiments. The OS1-64 configuration also demonstrated competitive performance, surpassing the VLP-16 configuration in most cases. However, these findings did not directly align with what we observed regarding the impact of individual LiDAR configuration parameters in Table \ref{tab:mavs-same-indiv-results} and Table \ref{tab:mavs-diff-indiv-results}, particularly the notion that increasing the vertical resolution consistently improved the model's performance in both scenarios. This suggests that the interplay between different parameters can lead to complex and sometimes unexpected impacts on the model's performance.


\subsection{Impact of Sensor Domain Shift between Training and Testing}
\label{sub-sec:Impact of Sensor Domain Shift between Training and Testing}
We assessed the impact of sensor domain shift between training and test by evaluating the Cylinder3D model's performance on MAVS simulated datasets generated with varying LiDAR configurations under sensor domain shift, considering both cases with and without spatial shift. Table \ref{tab:mavs-same-shifted-results} (Appendix \ref{sec:miscA}) presents the experimental results of the Cylinder3D model's performance on MAVS simulated datasets with varying LiDAR parameters under sensor domain shift and without spatial shift. As Table \ref{tab:mavs-same-shifted-results} shows, the model experienced a substantial performance decline when the training and testing datasets had different vertical resolution (vres) values. The most considerable performance drop occurred when the training data had a vres of 16, and the testing data had a vres of 64, yielding a mean mIoU of 81.48. Similarly, the model's performance significantly decreased when the training and testing datasets had different horizontal resolution (hres) values. The worst performance was observed when the model was trained on hres: 0.1 and tested on hres: 0.4, with a mean mIoU of 72.19.

Table \ref{tab:mavs-diff-shifted-results} (Appendix \ref{sec:miscA}) presents performance metrics for the Cylinder3D model on MAVS simulated datasets under sensor domain shifts and spatial shifts. The model's performance significantly decreased when there was a spatial shift between the training and testing datasets compared to when there was no spatial shift; the mean mIoU is below 40 for all conditions. In cases where the model was pre-trained on the original dataset, the most substantial performance degradation occurred when tested with a vres of 64. The worst performance degradation among all cases was observed when the model was trained on hres: 0.1 and tested on the dataset with hres: 0.4, resulting in a mean performance of 22.78, compared to a mean performance of 39.46 when trained and tested on hres: 0.1. In this scenario, the model's performance was susceptible to changes in vertical and horizontal resolution. When the training and testing datasets had different vertical resolution values, the model's performance dropped significantly, as shown in the experiments with vres: 16 and vres: 64. A similar trend was observed for horizontal resolution, with the poorest performance when the model was trained on hres: 0.1 and tested on hres: 0.4.

The results of the same experiments using MAVS simulated datasets generated with VLP-16, Velodyne Ultra Puck, and OS1 configurations are presented in Tables \ref{tab:mavs-same-shifted-combined-results} and \ref{tab:mavs-diff-shifted-combined-results} (Appendix \ref{sec:miscA}). Table \ref{tab:mavs-same-shifted-combined-results} shows the results under sensor domain shift without spatial domain shift. The performance slightly decreased when testing the model pre-trained on the Velodyne Ultra Puck dataset with data from other sensors, but it did not significantly impact the results. Similarly, when training with the OS1-64, the performance slightly decreased when tested with data from other sensors, but it also did not significantly impact the results. A significant performance drop compared to the other two sensor datasets occurred when there was a combination of sensor and spatial domain shifts was present. In line with the other experiments involving spatial domain shift, Table \ref{tab:mavs-diff-shifted-combined-results} shows that the model's performance significantly declines compared to the results from experiments without spatial domain shift.

Overall, the experiments indicated that training and testing on the same sensor data often result in better performance than when the training and testing datasets are from different sensors. Nevertheless, it is worth noting that the sensor domain shift between training and testing datasets did not always lead to a decline in performance. In some instances, better performance was achieved when training and testing on different sensor data rather than using the same sensor data. However, most experiments showed improved performance when the sensor data for training and testing were consistent, with some exceptions. The experiments revealed that performance degradation occurs in most scenarios when a sensor domain shift is present, especially when there is a spatial domain shift. Notably, performance decreased more significantly when testing on datasets with higher vertical resolution than the training datasets compared to when testing on datasets with the same vertical resolution. Conversely, when testing on datasets with lower vertical resolution than the training datasets, the performance varied depending on the presence of a spatial domain shift. Without spatial domain shift, performance decreased more when testing on datasets with lower vertical resolution than those with the same vertical resolution. However, with spatial domain shift, testing on datasets with lower vertical resolution actually led to improved performance compared to testing on datasets with the same vertical resolution.

As observed in sub-section \ref{sub-sec:Effect of Change in Individual LiDAR Parameter}, when there was no sensor domain shift or spatial domain shift, increasing the horizontal resolution of the training dataset improved the performance on the test dataset. However, when a spatial domain shift was present, performance decreased. On the other hand, without sensor domain shift, the performance was generally lower when testing on domains with different horizontal resolutions compared to testing on domains with the same horizontal resolution, regardless of the presence of spatial domain shift.

Experiments with the three sensors (Table \ref{tab:mavs-same-shifted-combined-results} and Table \ref{tab:mavs-diff-shifted-combined-results}) did not show results that fully aligned with the previous findings, where increase in vertical resolution led to performance improvement. However, in all other scenarios, the results were consistent with the findings obtained when testing with only one configuration change.


\subsection{Results on RELLIS-3D}
\label{sub-sec:Results on RELLIS-3D}

The experimental results using the RELLIS-3D dataset are presented in Table \ref{tab:rellis-same-results} and Table \ref{tab:rellis-diff-results}. These experiments were conducted to simulate practical scenarios in which LiDAR sensors might be replaced with other sensor models, allowing us to examine the effects of sensor and spatial domain shifts on model performance using a real-world off-road dataset. The results for the real off-road environment dataset RELLIS-3D displayed similar trends to the previously discussed MAVS simulation dataset results as shown in Tables \ref{tab:mavs-same-shifted-combined-results} and \ref{tab:mavs-diff-shifted-combined-results} (Appendix \ref{sec:miscA}), discussed in sub-section \ref{sub-sec:Effect of Change in Individual LiDAR Parameter}.

\begin{table}[h]
\caption{Results on RELLIS-3D under no spatial domain shift.}
\label{tab:rellis-same-results}
\begin{center}
\resizebox{\textwidth}{!}{%
\begin{tabular}{l|c|c|c|c|c|c|c|c|c|c|c|c|c|c|c}
\hline
\rule[-1ex]{0pt}{3.5ex} Trained on - Tested on & grass & tree & bush & concrete & mud & person & puddle & rubble & barrier & log & fence & vehicle & pole & water & mean \\ 
\hline
\hline
\rule[-1ex]{0pt}{3.5ex} Velodyne Ultra Puck - Velodyne Ultra Puck & 69.92 & 64.07 & 77.77 & 87.80 & 5.030 & 88.41  & 21.73  & 1.98   & 38.93   & 4.17 & 8.11  & 28.18   & 12.11 & 0.00     & 36.30 \\        
\rule[-1ex]{0pt}{3.5ex} Velodyne Ultra Puck - OS1-64 & 30.90  & 59.32 & 53.92 & 46.65 & 0.28 & 3.39   & 1.49   & 0.00    & 2.44  & 0.00  & 0.60   & 7.64    & 12.80  & 0.00    & 15.68 \\
\rule[-1ex]{0pt}{3.5ex} OS1-64 - Velodyne Ultra Puck & 0.64  & 2.86  & 0.96  & 0.00 & 0.28 & 27.64  & 0.00     & 0.00  & 0.00  & 0.00  & 0.00    & 0.24    & 0.00  & 0.00    & 2.32 \\
\rule[-1ex]{0pt}{3.5ex} OS1-64 - OS1-64 & 64.05 & 74.76 & 71.09 & 81.24 & 9.47 & 86.39  & 23.69  & 0.59   & 67.58   & 1.07 & 3.46  & 65.71  & 56.33 & 0.00  & 43.24 \\
\hline
\end{tabular}}
\end{center}
\end{table} 

\begin{table}[h]

\caption{Results on RELLIS-3D under spatial domain shift.}
\label{tab:rellis-diff-results}
\begin{center}
\resizebox{\textwidth}{!}{%
\begin{tabular}{l|c|c|c|c|c|c|c|c|c|c|c|c|c|c|c}
\hline
\rule[-1ex]{0pt}{3.5ex} Trained on - Tested on & grass & tree & bush & concrete & mud & person & puddle & rubble & barrier & log & fence & vehicle & pole & water & mean \\ 
\hline
\hline
\rule[-1ex]{0pt}{3.5ex} Velodyne Ultra Puck - Velodyne Ultra Puck & 44.51 & 72.85 & 68.49 & 44.09 & 4.85 & 86.36 & 12.10 & 52.66 & 77.27 & 3.08 & 0.00 & 27.23 & 0.00 & 0.00 & 35.25 \\ 
\rule[-1ex]{0pt}{3.5ex} Velodyne Ultra Puck - OS1-64 & 44.57 & 69.82 & 37.73 & 2.41     & 0.00     & 0.14   & 0.02   & 2.42   & 0.94    & 0.12  & 0.00     & 11.91   & 0.00   & 0.00     & 12.14 \\ 
\rule[-1ex]{0pt}{3.5ex} OS1-64 - Velodyne Ultra Puck & 0.34  & 3.05  & 14.39 & 0.05     & 0.02  & 0.61   & 0.20   & 0.00     & 0.00       & 0.00     & 0.01  & 0.22    & 0.14 & 0.00     & 1.36  \\ 
\rule[-1ex]{0pt}{3.5ex} OS1-64 - OS1-64 & 51.26 & 84.29 & 43.76 & 38.27    & 10.41 & 85.67  & 35.10   & 54.72  & 63.44   & 7.19  & 0.00     & 38.94   & 0.00   & 0.00     & 36.68 \\ 
\hline
\end{tabular}}
\end{center}
\end{table} 

However, there were some differences in sensor performance and the impact of sensor domain changes between the experiments on the simulated datasets and the RELLIS-3D dataset. In contrast to the scenarios trained and tested on the Velodyne Ultra Puck, which showed the best performance regardless of spatial domain changes in Table \ref{tab:mavs-same-shifted-combined-results} and Table \ref{tab:mavs-diff-shifted-combined-results}, both experiments with RELLIS-3D showed that the OS1-64 sensor model when trained and tested, achieved the best performance irrespective of spatial domain changes. Furthermore, while the pre-trained model on OS1-64 improved performance when tested on the Velodyne Ultra Puck dataset in the experiments on the simulated datasets, the RELLIS-3D experiment exhibited a significant performance decline.

Based on these findings, it can be inferred that, in the absence of sensor domain changes, the OS1-64 sensor model, with its higher horizontal and vertical resolution, can process more information and deliver better performance. However, when sensor domain changes are present, this advantage might lead to overfitting in a specific domain, resulting in a lack of generalization capability and a substantial decrease in performance when tested on a different sensor.

\label{sec:results}

\section{DISCUSSION}
In this section, we discuss the limitations of our study and the implications for future research. First, our simulated dataset created using MAVS may not accurately represent real-world scenarios due to its limited scope and diversity. Due to the limited number of classes in the datasets and the lack of variability in the objects included in the simulated off-road environments, these simplified environments may not fully capture the complexity of real-world 3D LiDAR semantic segmentation. Second, our study focused exclusively on the Cylinder3D model, raising the question of whether the results can be generalized to other 3D LiDAR semantic segmentation models. Our future research should examine the performance of various models and explore the potential benefits of combining multiple techniques to address this concern. Third, we identified a significant number of annotation errors in the RELLIS-3D dataset, which can be attributed to the manual labeling process. This issue highlights the challenges associated with human annotation, as it is nearly impossible to achieve the same level of accuracy as a simulation program like MAVS. Our future research should focus on developing self-supervised approaches for 3D LiDAR segmentation and methods for improving semantic segmentation models to address the problems caused by human manual annotation processes and the lack of diverse datasets for off-road 3D LiDAR semantic segmentation. The impact of sensor and spatial domain shifts, as well as configuration changes, on semantic segmentation performance was evident in our study. This underscores the importance of researching domain adaptation techniques to alleviate these issues and enhance the robustness of 3D LiDAR semantic segmentation models in diverse settings. In summary, our study has several limitations, including the reliance on simulated datasets, the exclusive use of the Cylinder3D model, and annotation errors in the RELLIS-3D dataset. These factors, coupled with the challenges posed by sensor and spatial domain shifts, suggest the need for further investigation and the development of novel approaches to improve the performance and applicability of 3D LiDAR semantic segmentation models in real-world situations.
\label{sec:discussion}

\section{CONCLUSION \& FUTURE WORK}
In summary, this study investigated and analyzed the effect of varying different LiDAR configurations such as number of channels, field of view, range, resolution, and sensor mounting position on the resulting point cloud semantic segmentation performance. Experiments were conducted using the Cylinder3D neural network with LiDAR data acquired from the RELLIS-3D dataset as well as the MAVS simulator. Results show that the number of channels had a significant effect on the segmentation accuracy whereas the field of view, range, resolution, and sensor position had minor effects on the segmentation accuracy. In addition, sensor domain shift (i.e. using different types of LiDARs between training and testing data) as well as spatial domain shift (i.e. having different types of vegetation and terrain between training and testing data) caused significant degradation in the segmentation accuracy. The results from this study suggests that the following recommendations can be made for deployment of LiDAR point cloud semantic segmentation models: (i) using LiDARs with larger number of channels, which results in larger vertical field of view and vertical resolution, is advantageous for obtaining more accurate models (ii) using different types of LiDARs for training and deployment should be avoided (iii) using training data from a different spatial domain compared to the deployment environment should be avoided.

In future work, we plan to expand the scope of the study by considering a larger number of different LiDAR models as well as a larger number of different environments. In addition, we will investigate the effect of changing LiDAR configuration on different types of neural network architectures such as Point Transformers and Point Pillars.
We will also explore more advanced learning techniques such as domain adaptation, self-supervised learning, or contrastive learning that can effectively make use of training data despite the domain shift between training and deployment.
\label{sec:conclusion}

\appendix    

\acknowledgments 
The work reported herein was supported by the National Science Foundation (NSF) (Award \#IIS-2153101). Any opinions, findings, conclusions or recommendations expressed in this material are those of the authors and do not necessarily reflect the views of the NSF.

\section{Full experimental result tables}
(Continued on the next page)

\begin{table}[ht]
\caption{Performance metrics for the Cylinder3D model on MAVS simulated datasets with varying individual LiDAR parameters under sensor domain shift and no spatial shift.}
\label{tab:mavs-same-shifted-results}
\begin{center}
\resizebox{\textwidth}{!}{%
\begin{tabular}{l|c|c|c|c|c|c}
\hline
\rule[-1ex]{0pt}{3.3ex} Trained on - Tested on & High Vegetation & Low Vegetation & Obstacle & Rough Trail & Smooth Trail & Mean \\
\hline
\hline
\rule[-1ex]{0pt}{3.3ex} \textbf{Trained on original} & & & & & & \\
\rule[-1ex]{0pt}{3.3ex} original - original & 99.31 & 90.36 & 92.46 & 98.86 & 92.12 & 94.62 \\
\rule[-1ex]{0pt}{3.3ex} 100m (original) - 150m & 98.99 & 87.23 & 86.76 & 98.22 & 90.73 & 92.38 \\
\rule[-1ex]{0pt}{3.3ex} 100m (original) - 200m & 98.93 & 85.86 & 86.01 & 97.95 & 89.36 & 91.62 \\
\rule[-1ex]{0pt}{3.3ex} [-15, 15] (original) - [-10, 20] & 99.07 & 88.24 & 86.40 & 98.06 & 86.83 & 91.72 \\
\rule[-1ex]{0pt}{3.3ex} [-15, 15] (original) - [-20, 10] & 98.63 & 89.68 & 87.96 & 98.61 & 93.58 & 93.69 \\
\rule[-1ex]{0pt}{3.3ex} (vres: 16) (original) - (vres: 32) & 99.11 & 90.55 & 89.57 & 98.27 & 89.18 & 93.34 \\
\rule[-1ex]{0pt}{3.3ex} (vres: 16) (original) - (vres: 64) & 98.06 & 78.17 & 57.86 & 90.58 & 82.77 & 81.48 \\
\rule[-1ex]{0pt}{3.3ex} (hres: 0.2)  (original) - (hres: 0.1) & 99.45 & 91.20 & 93.68 & 98.98 & 93.07 & 95.28 \\
\rule[-1ex]{0pt}{3.3ex} (hres: 0.2)  (original) - (hres: 0.4) & 98.96 & 84.53 & 87.09 & 96.80 & 73.12 & 88.10 \\
\rule[-1ex]{0pt}{3.3ex} (0m, 0m, 1.2m) (original) - 10cm up & 99.37 & 89.43 & 92.05 & 98.78 & 90.64 & 94.05 \\
\rule[-1ex]{0pt}{3.3ex} (0m, 0m, 1.2m) (original) - 10cm down & 99.19 & 90.87 & 92.44 & 98.85 & 93.06 & 94.88 \\
\rule[-1ex]{0pt}{3.3ex} (0m, 0m, 1.2m) (original) - 10cm left & 99.31 & 90.32 & 92.45 & 98.85 & 92.08 & 94.60 \\
\rule[-1ex]{0pt}{3.3ex} (0m, 0m, 1.2m) (original) - 10cm right & 99.31 & 90.29 & 92.42 & 98.84 & 91.96 & 94.56 \\
\hline
\rule[-1ex]{0pt}{3.3ex} \textbf{Trained on 150m} & & & & & & \\
\rule[-1ex]{0pt}{3.3ex} 150m - 150 m & 99.11 & 90.91 & 90.28 & 98.74 & 93.17 & 94.44 \\
\rule[-1ex]{0pt}{3.3ex} 150m - 100m (original) & 99.17 & 88.45 & 92.51 & 99.07 & 93.17 & 94.48 \\
\rule[-1ex]{0pt}{3.3ex} 150m - 200 m & 99.09 & 90.07 & 90.12 & 98.44 & 93.00 & 94.15 \\
\hline
\rule[-1ex]{0pt}{3.3ex} \textbf{Trained on 200m} & & & & & & \\
\rule[-1ex]{0pt}{3.3ex} 200m - 200 m & 99.14 & 90.83 & 90.32 & 98.56 & 92.77 & 94.32 \\
\rule[-1ex]{0pt}{3.3ex} 200m - 100m (original) & 99.21 & 89.41 & 92.46 & 99.02 & 92.64 & 94.54 \\
\rule[-1ex]{0pt}{3.3ex} 200m - 150 m & 99.14 & 91.42 & 90.12 & 98.73 & 92.83 & 94.44 \\
\hline
\rule[-1ex]{0pt}{3.3ex} \textbf{Trained on vertical FOV [-10, 20]} & & & & & & \\
\rule[-1ex]{0pt}{3.3ex} [-10, 20] - [-10, 20] & 99.21 & 91.47 & 88.87 & 98.32 & 87.71 & 93.12 \\
\rule[-1ex]{0pt}{3.3ex} [-10, 20] - [-15, 15] (original) & 99.22 & 88.88 & 90.70 & 98.19 & 86.11 & 92.62 \\
\rule[-1ex]{0pt}{3.3ex} [-10, 20] - [-20, 10] & 98.75 & 87.58 & 78.69 & 98.49 & 89.13 & 90.528 \\
\hline
\rule[-1ex]{0pt}{3.3ex} \textbf{Trained on vertical FOV [-20, 10]} & & & & & & \\
\rule[-1ex]{0pt}{3.3ex} [-20, 10] - [-20, 10] & 98.83 & 92.48 & 90.12 & 98.76 & 94.18 & 94.87 \\
\rule[-1ex]{0pt}{3.3ex} [-20, 10] - [-15, 15] (original) & 99.27 & 90.14 & 91.64 & 98.92 & 92.73 & 94.54 \\
\rule[-1ex]{0pt}{3.3ex} [-20, 10] - [-10, 20] & 99.2 & 91.45 & 88.47 & 98.38 & 89.24 & 93.34 \\
\hline
\rule[-1ex]{0pt}{3.3ex} \textbf{Trained on (vres: 32)} & & & & & & \\
\rule[-1ex]{0pt}{3.3ex} (vres: 32) - (vres: 32) & 99.54 & 94.70 & 93.98 & 98.98 & 93.98 & 96.24 \\
\rule[-1ex]{0pt}{3.3ex} (vres: 32) - (vres: 16) (original) & 99.28 & 88.12 & 91.42 & 98.72 & 91.05 & 93.72 \\
\rule[-1ex]{0pt}{3.3ex} (vres: 32) - (vres: 64) & 99.57 & 94.10 & 94.20 & 98.80 & 93.98 & 96.133 \\
\hline
\rule[-1ex]{0pt}{3.3ex} \textbf{Trained on (vres: 64)} & & & & & & \\
\rule[-1ex]{0pt}{3.3ex} (vres: 64) - (vres: 64) & 99.69 & 95.15 & 95.37 & 98.99 & 95.20 & 96.88 \\
\rule[-1ex]{0pt}{3.3ex} (vres: 64) - (vres: 16) (original) & 98.57 & 80.99 & 87.84 & 98.23 & 89.06 & 90.94 \\
\rule[-1ex]{0pt}{3.3ex} (vres: 64) - (vres: 32) & 99.48 & 94.51 & 93.02 & 98.90 & 93.40 & 95.86 \\
\hline
\rule[-1ex]{0pt}{3.3ex} \textbf{Trained on (hres: 0.1)} & & & & & & \\
\rule[-1ex]{0pt}{3.3ex} (hres: 0.1) - (hres: 0.1) & 99.44 & 92.61 & 94.51 & 99.21 & 94.42 & 96.04 \\
\rule[-1ex]{0pt}{3.3ex} (hres: 0.1) - (hres: 0.2) (original) & 99.15 & 88.22 & 91.27 & 98.24 & 85.63 & 92.50 \\
\rule[-1ex]{0pt}{3.3ex} (hres: 0.1) - (hres: 0.4) & 98.53 & 61.06 & 79.67 & 94.66 & 27.01 & 72.19 \\
\hline
\rule[-1ex]{0pt}{3.3ex} \textbf{Trained on (hres: 0.4)} & & & & & & \\
\rule[-1ex]{0pt}{3.3ex} (hres: 0.4) - (hres: 0.4) & 99.00 & 87.14 & 88.71 & 98.77 & 91.73 & 93.07 \\
\rule[-1ex]{0pt}{3.3ex} (hres: 0.4) - (hres: 0.2) (original) & 99.05 & 87.65 & 87.84 & 98.48 & 89.55 & 92.52 \\
\rule[-1ex]{0pt}{3.3ex} (hres: 0.4) - (hres: 0.1) & 98.66 & 80.28 & 80.4 & 97.85 & 84.87 & 88.41 \\
\hline
\end{tabular}}
\end{center}
\end{table}

\begin{table}
\caption{Performance metrics for the Cylinder3D model on MAVS simulated datasets with varying individual LiDAR parameters under sensor domain shift and spatial shift.}
\label{tab:mavs-diff-shifted-results}
\begin{center}
\resizebox{\textwidth}{!}{%
\begin{tabular}{l|c|c|c|c|c|c}
\hline
\rule[-1ex]{0pt}{3.3ex} Trained on - Tested on & High Vegetation & Low Vegetation & Obstacle & Rough Trail & Smooth Trail & Mean \\
\hline
\hline
\rule[-1ex]{0pt}{3.3ex} \textbf{Trained on original} & & & & & & \\
\rule[-1ex]{0pt}{3.3ex} original - original & 47.10 & 11.92 & 8.42  & 43.88 & 81.68 & 38.60 \\
\rule[-1ex]{0pt}{3.3ex} 100m (original) - 150m & 37.21 & 5.28  & 8.59  & 14.36 & 79.97 & 29.08 \\
\rule[-1ex]{0pt}{3.3ex} 100m (original) - 200m & 37.25 & 5.26  & 8.59  & 14.30 & 79.76 & 29.03 \\
\rule[-1ex]{0pt}{3.3ex} [-15, 15] (original) - [-10, 20] & 48.46 & 5.26  & 7.12  & 8.70  & 71.46 & 28.20 \\
\rule[-1ex]{0pt}{3.3ex} [-15, 15] (original) - [-20, 10] & 26.94 & 6.40  & 6.27  & 24.37 & 82.04 & 29.20 \\
\rule[-1ex]{0pt}{3.3ex} (vres: 16) - (vres: 32) & 37.52 & 3.47  & 5.69  & 19.76 & 73.50 & 27.99 \\
\rule[-1ex]{0pt}{3.3ex} (vres: 16) - (vres: 64) & 40.26 & 2.26  & 2.91  & 13.32 & 70.40 & 25.83 \\
\rule[-1ex]{0pt}{3.3ex} (hres: 0.2) - (hres: 0.1) & 44.74 & 7.18  & 9.34  & 39.41 & 82.68 & 36.67 \\
\rule[-1ex]{0pt}{3.3ex} (hres: 0.2) - (hres: 0.4) & 48.82 & 17.62 & 6.53  & 41.06 & 63.18 & 35.44 \\
\rule[-1ex]{0pt}{3.3ex} (0m, 0m, 1.2m) (original) - 10cm up & 53.39 & 14.19 & 7.34  & 48.23 & 81.46 & 40.92 \\
\rule[-1ex]{0pt}{3.3ex} (0m, 0m, 1.2m) (original) - 10cm down & 40.84 & 9.82  & 9.36  & 36.60 & 78.61 & 35.05 \\
\rule[-1ex]{0pt}{3.3ex} (0m, 0m, 1.2m) (original) - 10cm left & 47.08 & 11.94 & 8.40  & 43.72 & 81.64 & 38.56 \\
\rule[-1ex]{0pt}{3.3ex} (0m, 0m, 1.2m) (original) - 10cm right & 47.06 & 11.87 & 8.35  & 43.99 & 81.60 & 38.57 \\
\hline
\rule[-1ex]{0pt}{3.3ex} \textbf{Trained on 150m} & & & & & & \\
\rule[-1ex]{0pt}{3.3ex} 150m - 150 m  & 48.62 & 13.95 & 3.48  & 49.51 & 78.94 & 38.90 \\
\rule[-1ex]{0pt}{3.3ex} 150m - 100m (original)  & 61.68 & 20.00 & 5.35  & 66.42 & 81.15 & 46.92 \\
\rule[-1ex]{0pt}{3.3ex} 150m - 200 m & 48.63 & 13.93 & 3.49  & 49.34 & 78.92 & 38.86 \\
\hline
\rule[-1ex]{0pt}{3.3ex} \textbf{Trained on 200m} & & & & & & \\
\rule[-1ex]{0pt}{3.3ex} 200m - 200 m & 59.20 & 13.90 & 2.17  & 55.15 & 79.72 & 42.03 \\
\rule[-1ex]{0pt}{3.3ex} 200m - 100m (original) & 64.57 & 19.61 & 5.10  & 70.00 & 82.69 & 48.40 \\
\rule[-1ex]{0pt}{3.3ex} 200m - 150 m & 59.27 & 13.91 & 2.17  & 55.25 & 80.38 & 42.20 \\
\hline
\rule[-1ex]{0pt}{3.3ex} \textbf{Trained on vertical FoV [-10, 20]} & & & & & & \\
\rule[-1ex]{0pt}{3.3ex} [-10, 20] - [-10, 20] & 48.91 & 8.18  & 11.34 & 6.85  & 52.77 & 25.61 \\
\rule[-1ex]{0pt}{3.3ex} [-10, 20] - [-15, 15] (original) & 43.75 & 12.49 & 14.10 & 32.41 & 75.58 & 35.67 \\
\rule[-1ex]{0pt}{3.3ex} [-10, 20] - [-20, 10] & 26.96 & 6.56  & 7.77  & 31.08 & 78.17 & 30.11 \\
\hline
\rule[-1ex]{0pt}{3.3ex} \textbf{Trained on vertical FoV [-20, 10]} & & & & & & \\
\rule[-1ex]{0pt}{3.3ex} [-20, 10] - [-20, 10] & 41.58 & 10.29 & 2.67  & 67.91 & 84.67 & 41.42 \\
\rule[-1ex]{0pt}{3.3ex} [-20, 10] - [-15, 15] (original) & 58.59 & 17.58 & 5.95 & 68.44 & 82.17 & 46.546 \\
\rule[-1ex]{0pt}{3.3ex} [-20, 10] - [-10, 20] & 57.38 & 9.43  & 2.96  & 38.44 & 71.47 & 35.94 \\
\hline
\rule[-1ex]{0pt}{3.3ex} \textbf{Trained on (vres: 32)} & & & & & & \\
\rule[-1ex]{0pt}{3.3ex} (vres: 32) - (vres: 32) & 50.62 & 7.48  & 2.88  & 62.61 & 87.22 & 42.16 \\
\rule[-1ex]{0pt}{3.3ex} (vres: 32) - (vres: 16) (original) & 66.63 & 16.03 & 5.13  & 69.34 & 85.28 & 48.48 \\
\rule[-1ex]{0pt}{3.3ex} (vres: 32) - (vres: 64) & 47.43 & 5.62  & 3.01  & 56.34 & 86.04 & 39.69 \\
\hline
\rule[-1ex]{0pt}{3.3ex} \textbf{Trained on (vres: 64)} & & & & & & \\
\rule[-1ex]{0pt}{3.3ex} (vres: 64) - (vres: 64) & 48.99 & 4.91  & 2.88  & 68.74 & 85.38 & 42.18 \\
\rule[-1ex]{0pt}{3.3ex} (vres: 64) - (vres: 16) (original) & 55.27 & 19.81 & 9.61  & 69.31 & 78.61 & 46.52 \\
\rule[-1ex]{0pt}{3.3ex} (vres: 64) - (vres: 32) & 48.31 & 7.27  & 3.22  & 66.24 & 84.15 & 41.84 \\
\hline
\rule[-1ex]{0pt}{3.3ex} \textbf{Trained on (hres: 0.1)} & & & & & & \\
\rule[-1ex]{0pt}{3.3ex} (hres: 0.1) - (hres: 0.1) & 46.65 & 10.93 & 9.14  & 49.70 & 80.89 & 39.46 \\
\rule[-1ex]{0pt}{3.3ex} (hres: 0.1) - (hres: 0.2) (original) & 47.67 & 13.90 & 7.23  & 44.64 & 68.24 & 36.34 \\
\rule[-1ex]{0pt}{3.3ex} (hres: 0.1) - (hres: 0.4) & 45.35 & 16.74 & 5.68  & 30.11 & 16.01 & 22.78 \\
\hline
\rule[-1ex]{0pt}{3.3ex} \textbf{Trained on (hres: 0.4)} & & & & & & \\
\rule[-1ex]{0pt}{3.3ex} (hres: 0.4) - (hres: 0.4) & 57.96 & 11.81 & 4.93  & 60.11 & 71.29 & 41.22 \\
\rule[-1ex]{0pt}{3.3ex} (hres: 0.4) - (hres: 0.2) (original) & 53.62 & 6.96  & 5.31  & 61.58 & 71.29 & 39.75 \\
\rule[-1ex]{0pt}{3.3ex} (hres: 0.4) - (hres: 0.1) & 50.25 & 3.84  & 5.49  & 60.31 & 66.66 & 37.31 \\
\hline
\end{tabular}}
\end{center}
\end{table} 

\begin{table}
\caption{Results of the experiments on the MAVS simulated datasets generated with VLP-16, Velodyne Ultra Puck, and OS1 configurations under sensor domain shift but no spatial domain shift.}
\label{tab:mavs-same-shifted-combined-results}
\begin{center}
\resizebox{\textwidth}{!}{%
\begin{tabular}{l|c|c|c|c|c|c}
\hline
\rule[-1ex]{0pt}{3.3ex} Trained on - Tested on & High Vegetation & Low Vegetation & Obstacle & Rough Trail & Smooth Trail & Mean \\
\hline
\hline
\rule[-1ex]{0pt}{3.3ex} \textbf{Trained on VLP-16} & & & & & & \\
\rule[-1ex]{0pt}{3.3ex} VLP-16 - VLP-16 & 99.31 & 90.36 & 92.46 & 98.86 & 92.12 & 94.62 \\
\rule[-1ex]{0pt}{3.3ex} VLP-16 - Velodyne Ultra Puck & 99.00 & 80.59 & 64.35 & 95.10 & 93.69 & 86.55 \\
\rule[-1ex]{0pt}{3.3ex} VLP-16 - OS1-64 & 97.23 & 75.28 & 59.28 & 87.79 & 63.93 & 76.70 \\
\hline
\rule[-1ex]{0pt}{3.3ex} \textbf{Trained on Velodyne Ultra Puck} & & & & & & \\
\rule[-1ex]{0pt}{3.3ex} Velodyne Ultra Puck - Velodyne Ultra Puck & 99.45 & 94.05 & 92.94 & 98.94 & 95.86 & 96.25 \\
\rule[-1ex]{0pt}{3.3ex} Velodyne Ultra Puck - VLP-16 & 99.15 & 88.23 & 91.57 & 98.82 & 91.99 & 93.95 \\
\rule[-1ex]{0pt}{3.3ex} Velodyne Ultra Puck - OS1-64 & 99.19 & 91.11 & 89.57 & 97.55 & 84.40 & 92.37 \\
\hline
\rule[-1ex]{0pt}{3.3ex} \textbf{Trained on OS1-64} & & & & & & \\
\rule[-1ex]{0pt}{3.3ex} OS1-64 & 99.48 & 93.18 & 93.27 & 98.80 & 94.33 & 95.81 \\
\rule[-1ex]{0pt}{3.3ex} OS1-64 - VLP-16 & 98.48 & 84.99 & 85.40 & 98.18 & 87.90 & 90.99 \\
\rule[-1ex]{0pt}{3.3ex} OS1-64 - Velodyne Ultra Puck & 98.90 & 92.13 & 86.99 & 97.65 & 89.53 & 93.04 \\
\hline
\end{tabular}}
\end{center}
\end{table} 

\begin{table}[h]
\caption{Results of the experiments on the MAVS simulated datasets generated with VLP-16, Velodyne Ultra Puck, and OS1 configurations under both sensor domain shift and spatial domain shift.}
\label{tab:mavs-diff-shifted-combined-results}
\begin{center}
\resizebox{\textwidth}{!}{%
\begin{tabular}{l|c|c|c|c|c|c}
\hline
\rule[-1ex]{0pt}{3.3ex} Trained on - Tested on & High Vegetation & Low Vegetation & Obstacle & Rough Trail & Smooth Trail & Mean \\
\hline
\hline
\rule[-1ex]{0pt}{3.3ex} \textbf{Trained on VLP-16} & & & & & & \\
\rule[-1ex]{0pt}{3.3ex} VLP-16 - VLP-16  & 47.10 & 11.92 & 8.42 & 43.88 & 81.68 & 38.60 \\
\rule[-1ex]{0pt}{3.3ex} VLP-16 - Velodyne Ultra Puck & 39.48 & 5.14  & 2.41 & 36.43 & 80.19 & 32.73 \\
\rule[-1ex]{0pt}{3.3ex} VLP-16 - OS1-64 & 39.23 & 3.57  & 3.80 & 10.75 & 57.29 & 22.93 \\
\hline
\rule[-1ex]{0pt}{3.3ex} \textbf{Trained on Velodyne Ultra Puck} & & & & & & \\
\rule[-1ex]{0pt}{3.3ex} Velodyne Ultra Puck - Velodyne Ultra Puck & 56.39 & 7.90  & 2.06 & 75.59 & 88.88 & 46.16 \\
\rule[-1ex]{0pt}{3.3ex} Velodyne Ultra Puck - VLP-16 & 61.89 & 15.99 & 4.96 & 66.39 & 81.76 & 46.20 \\
\rule[-1ex]{0pt}{3.3ex} Velodyne Ultra Puck - OS1-64 & 59.54 & 7.43  & 1.89 & 45.06 & 76.70 & 38.13 \\
\hline
\rule[-1ex]{0pt}{3.3ex} \textbf{Trained on OS1-64} & & & & & & \\
\rule[-1ex]{0pt}{3.3ex} OS1-64 & 52.88 & 6.89  & 2.87 & 71.49 & 87.48 & 44.32 \\
\rule[-1ex]{0pt}{3.3ex} OS1-64 - VLP-16 & 53.32 & 16.99 & 8.89 & 72.46 & 79.96 & 46.32 \\
\rule[-1ex]{0pt}{3.3ex} OS1-64 - Velodyne Ultra Puck & 49.73 & 7.70  & 2.55 & 75.82 & 86.55 & 44.47 \\
\hline
\end{tabular}}
\end{center}
\end{table} 
\label{sec:miscA}

\clearpage
\section{RELLIS-3D Class distribution graphs}

\begin{figure}[h!]
   \begin{center}
   \resizebox{\textwidth}{!}{%
   \begin{tabular}{c} 
   \includegraphics{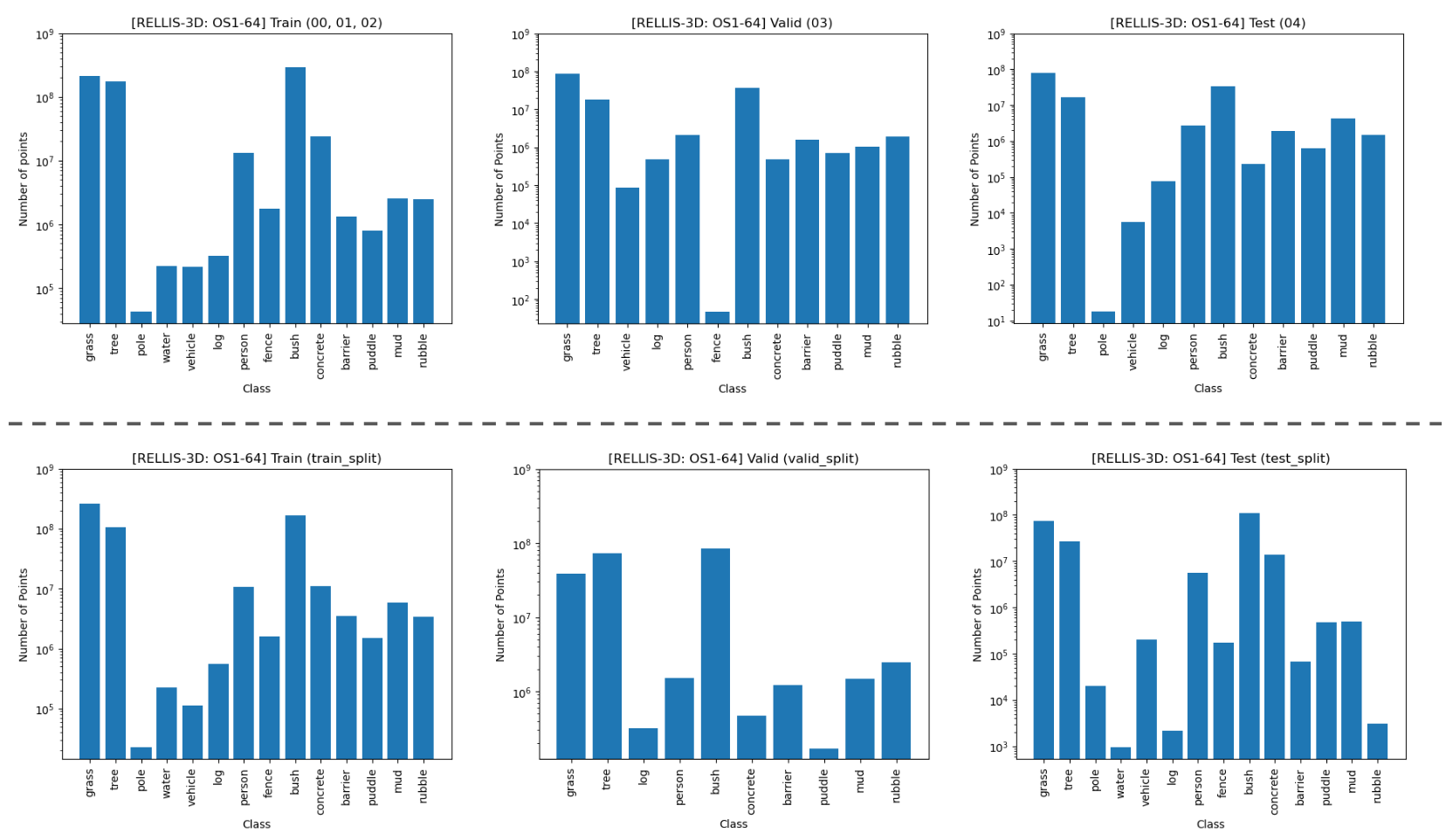}
	\end{tabular}}
	\end{center}
   \caption[example] 
   { \label{fig:rellis-os-dist} 
RELLIS-3D OS1-64 dataset's point cloud class distribution with respect to each split approaches in log scale. The top row shows distributions from the second approach (modified split), and the bottom row shows distributions from the first approach (original split).}
   \end{figure} 

\begin{figure}[h!]
   \begin{center}
   \resizebox{\textwidth}{!}{%
   \begin{tabular}{c} 
   \includegraphics{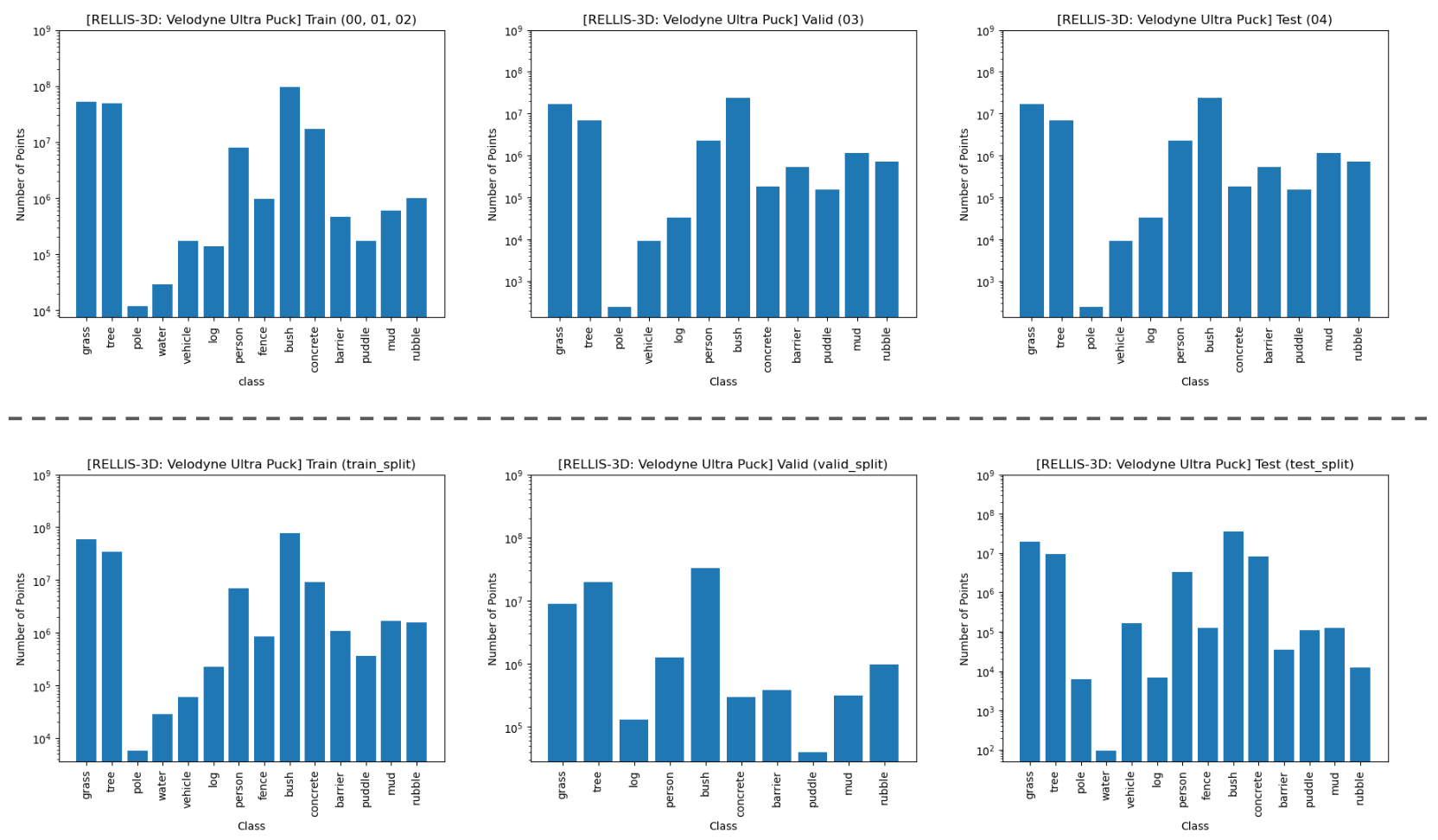}
	\end{tabular}}
	\end{center}
   \caption[example] 
   { \label{fig:rellis-vel-dist} 
RELLIS-3D Velodyne Ultra Puck dataset's point cloud class distribution with respect to each split approaches in log scale. The top row shows distributions from the second approach (modified split), and the bottom row shows distributions from the first approach (original split).}
   \end{figure}

\label{sec:miscB}

\clearpage
\bibliography{report} 

\begin{thebibliography}{10}

\bibitem{carballo2020}
Carballo, A., Lambert, J., Monrroy, A., Wong, D., Narksri, P., Kitsukawa, Y.,
  Takeuchi, E., Kato, S., and Takeda, K., ``Libre: The multiple 3d lidar
  dataset,'' in [{\em 2020 IEEE Intelligent Vehicles Symposium
  (IV)}{\nolinebreak\hspace{0.1em}]},   1094--1101 (2020).

\bibitem{9551643}
Viswanath, K., Singh, K., Jiang, P., Sujit, P., and Saripalli, S., ``Offseg: A
  semantic segmentation framework for off-road driving,'' in [{\em 2021 IEEE
  17th International Conference on Automation Science and Engineering
  (CASE)}{\nolinebreak\hspace{0.1em}]},   354--359 (2021).

\bibitem{kim2018crc}
Kim, P., Chen, J., and Cho, Y.~K., ``Autonomous mobile robot localization and
  mapping for unknown construction environments,'' in [{\em Construction
  Research Congress 2018}{\nolinebreak\hspace{0.1em}]},   147--156 (2018).

\bibitem{chen2022aei}
Chen, J. and Cho, Y.~K., ``Crackembed: Point feature embedding for crack
  segmentation from disaster site point clouds with anomaly detection,'' {\em
  Advanced Engineering Informatics}~{\bf 52},  101550 (2022).

\bibitem{orfd2022}
Min, C., Jiang, W., Zhao, D., Xu, J., Xiao, L., Nie, Y., and Dai, B., ``Orfd: A
  dataset and benchmark for off-road freespace detection,'' in [{\em 2022
  International Conference on Robotics and Automation
  (ICRA)}{\nolinebreak\hspace{0.1em}]},   2532--2538 (2022).

\bibitem{rs15010027}
Zhong, C., Li, B., and Wu, T., ``Off-road drivable area detection: A
  learning-based approach exploiting lidar reflection texture information,''
  {\em Remote Sensing}~{\bf 15}(1) (2023).

\bibitem{behley2019semantickitti}
Behley, J., Garbade, M., Milioto, A., Quenzel, J., Behnke, S., Stachniss, C.,
  and Gall, J., ``Semantickitti: A dataset for semantic scene understanding of
  lidar sequences,'' (2019).

\bibitem{caesar2020nuscenes}
Caesar, H., Bankiti, V., Lang, A.~H., Vora, S., Liong, V.~E., Xu, Q., Krishnan,
  A., Pan, Y., Baldan, G., and Beijbom, O., ``nuscenes: A multimodal dataset
  for autonomous driving,'' (2020).

\bibitem{pan2020semanticposs}
Pan, Y., Gao, B., Mei, J., Geng, S., Li, C., and Zhao, H., ``Semanticposs: A
  point cloud dataset with large quantity of dynamic instances,'' (2020).

\bibitem{jiang2020rellis3d}
Jiang, P., Osteen, P., Wigness, M., and Saripalli, S., ``Rellis-3d dataset:
  Data, benchmarks and analysis,'' (2020).

\bibitem{hudson2020mavs}
Hudson, C., Goodin, C., Miller, Z., Wheeler, W., and Carruth, D., ``Mississippi
  state university autonomous vehicle simulation library,'' in [{\em
  Proceedings of the Ground Vehicle Systems Engineering and Technology
  Symposium}{\nolinebreak\hspace{0.1em}]},   11--13 (2020).

\bibitem{zhu2021cylindrical}
Zhu, X., Zhou, H., Wang, T., Hong, F., Ma, Y., Li, W., Li, H., and Lin, D.,
  ``Cylindrical and asymmetrical 3d convolution networks for lidar
  segmentation,'' in [{\em Proceedings of the IEEE/CVF conference on computer
  vision and pattern recognition}{\nolinebreak\hspace{0.1em}]},   9939--9948
  (2021).

\bibitem{royo2019}
Royo, S. and Ballesta-Garcia, M., ``An overview of lidar imaging systems for
  autonomous vehicles,'' {\em Applied Sciences}~{\bf 9}(19) (2019).

\bibitem{chen2016slam}
Chen, J. and Cho, Y., ``Real-time 3d mobile mapping for the built
  environment,'' in [{\em 33rd International Symposium on Automation and
  Robotics in Construction (ISARC)}{\nolinebreak\hspace{0.1em}]},  (July 2016).

\bibitem{chen2017pad}
Chen, J., Fang, Y., Cho, Y.~K., and Kim, C., ``Principal axes descriptor for
  automated construction-equipment classification from point clouds,'' {\em
  Journal of Computing in Civil Engineering}~{\bf 31}(2),  04016058 (2017).

\bibitem{chen2021ral}
{Chen}, J., {Kira}, Z., and {Cho}, Y.~K., ``Lrgnet: Learnable region growing
  for class-agnostic point cloud segmentation,'' {\em IEEE Robotics and
  Automation Letters}~{\bf 6}(2),  2799--2806 (2021).

\bibitem{hirose2018gonet}
Hirose, N., Sadeghian, A., V{\'a}zquez, M., Goebel, P., and Savarese, S.,
  ``Gonet: A semi-supervised deep learning approach for traversability
  estimation,'' in [{\em 2018 IEEE/RSJ International Conference on Intelligent
  Robots and Systems (IROS)}{\nolinebreak\hspace{0.1em}]},   3044--3051, IEEE
  (2018).

\bibitem{chen2018ur}
{Chen}, J., {Kim}, P., {Cho}, Y.~K., and {Ueda}, J., ``Object-sensitive
  potential fields for mobile robot navigation and mapping in indoor
  environments,'' in [{\em 2018 15th International Conference on Ubiquitous
  Robots (UR)}{\nolinebreak\hspace{0.1em}]},   328--333 (June 2018).

\bibitem{price2020}
Price, L.~C., Chen, J., and Cho, Y.~K., ``Dynamic crane workspace update for
  collision avoidance during blind lift operations,'' in [{\em Proceedings of
  the 18th International Conference on Computing in Civil and Building
  Engineering}{\nolinebreak\hspace{0.1em}]},  Toledo~Santos, E. and Scheer, S.,
  eds.,  959--970, Springer International Publishing, Cham (2020).

\bibitem{Hu_2022_CVPR}
Hu, H., Liu, Z., Chitlangia, S., Agnihotri, A., and Zhao, D., ``Investigating
  the impact of multi-lidar placement on object detection for autonomous
  driving,'' in [{\em Proceedings of the IEEE/CVF Conference on Computer Vision
  and Pattern Recognition (CVPR)}{\nolinebreak\hspace{0.1em}]},   2550--2559
  (June 2022).

\bibitem{kim2017}
Kim, J., Jeong, J., Shin, Y.-S., Cho, Y., Roh, H., and Kim, A., ``Lidar
  configuration comparison for urban mapping system,'' in [{\em 2017 14th
  International Conference on Ubiquitous Robots and Ambient Intelligence
  (URAI)}{\nolinebreak\hspace{0.1em}]},   854--857 (2017).

\bibitem{zhang2021}
Zhang, W., Liu, N., and Zhang, Y., ``Learn to navigate maplessly with varied
  lidar configurations: A support point-based approach,'' {\em IEEE Robotics
  and Automation Letters}~{\bf 6}(2),  1918--1925 (2021).

\bibitem{mou2018}
Mou, S., Chang, Y., Wang, W., and Zhao, D., ``An optimal lidar configuration
  approach for self-driving cars,'' {\em CoRR}~{\bf abs/1805.07843} (2018).

\bibitem{qi2016pointnet}
Qi, C.~R., Su, H., Mo, K., and Guibas, L.~J., ``Pointnet: Deep learning on
  point sets for 3d classification and segmentation,'' {\em arXiv preprint
  arXiv:1612.00593}  (2016).

\bibitem{chen2018icra}
{Chen}, J., {Cho}, Y.~K., and {Ueda}, J., ``Sampled-point network for
  classification of deformed building element point clouds,'' in [{\em 2018
  IEEE International Conference on Robotics and Automation
  (ICRA)}{\nolinebreak\hspace{0.1em}]},   2164--2169 (2018).

\bibitem{Hu_2021_ICCV}
Hu, Z., Bai, X., Shang, J., Zhang, R., Dong, J., Wang, X., Sun, G., Fu, H., and
  Tai, C.-L., ``Vmnet: Voxel-mesh network for geodesic-aware 3d semantic
  segmentation,'' in [{\em Proceedings of the IEEE/CVF International Conference
  on Computer Vision (ICCV)}{\nolinebreak\hspace{0.1em}]},   15488--15498
  (October 2021).

\bibitem{Zhao_2021_ICCV}
Zhao, H., Jiang, L., Jia, J., Torr, P.~H., and Koltun, V., ``Point
  transformer,'' in [{\em Proceedings of the IEEE/CVF International Conference
  on Computer Vision (ICCV)}{\nolinebreak\hspace{0.1em}]},   16259--16268
  (October 2021).

\bibitem{chen2019ral}
{Chen}, J., {Cho}, Y.~K., and {Kira}, Z., ``Multi-view incremental segmentation
  of 3-d point clouds for mobile robots,'' {\em IEEE Robotics and Automation
  Letters}~{\bf 4}(2),  1240--1246 (2019).

\bibitem{yajima2021isarc}
Yajima, Y., Kim, S., Chen, J.~D., and Cho, Y., ``Fast online incremental
  segmentation of 3d point clouds from disaster sites,'' in [{\em Proceedings
  of the 38th International Symposium on Automation and Robotics in
  Construction (ISARC)}{\nolinebreak\hspace{0.1em}]},   341--348, International
  Association for Automation and Robotics in Construction (IAARC), Dubai, UAE
  (November 2021).

\bibitem{cat2022}
Sharma, S., Dabbiru, L., Hannis, T., Mason, G., Carruth, D.~W., Doude, M.,
  Goodin, C., Hudson, C., Ozier, S., Ball, J.~E., and Tang, B., ``Cat: Cavs
  traversability dataset for off-road autonomous driving,'' {\em IEEE
  Access}~{\bf 10},  24759--24768 (2022).

\bibitem{dabbiru2020}
Dabbiru, L., Goodin, C., Scherrer, N., and Carruth, D., ``Lidar data
  segmentation in off-road environment using convolutional neural networks
  (cnn),'' {\em SAE International Journal of Advances and Current Practices in
  Mobility}~{\bf 2},  3288--3292 (apr 2020).

\bibitem{Manivasagam_2020_CVPR}
Manivasagam, S., Wang, S., Wong, K., Zeng, W., Sazanovich, M., Tan, S., Yang,
  B., Ma, W.-C., and Urtasun, R., ``Lidarsim: Realistic lidar simulation by
  leveraging the real world,'' in [{\em Proceedings of the IEEE/CVF Conference
  on Computer Vision and Pattern Recognition
  (CVPR)}{\nolinebreak\hspace{0.1em}]},  (June 2020).

\bibitem{deschaud2021}
Deschaud, J., ``{KITTI-CARLA:} a kitti-like dataset generated by {CARLA}
  simulator,'' {\em CoRR}~{\bf abs/2109.00892} (2021).

\end{thebibliography}
\bibliographystyle{spiebib} 

\end{document}